\documentclass[sigconf, nonacm]{acmart}

\usepackage{multirow}
\usepackage{booktabs}
\usepackage{subcaption}
\usepackage{xspace}
\newcommand{\modelname}{AniXplore\xspace}
\usepackage[font=small]{caption}
\usepackage{float}
\usepackage{hyperref}
\hypersetup{
    colorlinks=true,
    linkcolor=blue,
    citecolor=blue,
    urlcolor=blue,
    pdfstartview=Fit,
    breaklinks=true
}
\renewcommand\footnotetextcopyrightpermission[1]{}

\begin{document}

\title[]{AnimeDL-2M: Million-Scale AI-Generated Anime Image Detection and Localization in Diffusion Era}

\author{Chenyang Zhu$^{1,2}$, Xing Zhang$^{2}$, Yuyang Sun$^{1,2}$, Ching-Chun Chang$^{2}$, Isao Echizen$^{1,2}$\\
$^{1}$The University of Tokyo, $^{2}$National Institute of Informatics, Japan\\}

\begin{abstract}
Recent advances in image generation, particularly diffusion models, have significantly lowered the barrier for creating sophisticated forgeries, making image manipulation detection and localization (IMDL) increasingly challenging. While prior work in IMDL has focused largely on natural images, the anime domain remains underexplored—despite its growing vulnerability to AI-generated forgeries. Misrepresentations of AI-generated images as hand-drawn artwork, copyright violations, and inappropriate content modifications pose serious threats to the anime community and industry. To address this gap, we propose \textbf{AnimeDL-2M}, the first large-scale benchmark for anime IMDL with comprehensive annotations. It comprises over two million images including real, partially manipulated, and fully AI-generated samples. Experiments indicate that models trained on existing IMDL datasets of natural images perform poorly when applied to anime images, highlighting a clear domain gap between anime and natural images. To better handle IMDL tasks in anime domain, we further propose \textbf{\modelname}, a novel model tailored to the visual characteristics of anime imagery. Extensive evaluations demonstrate that AniXplore achieves superior performance compared to existing methods. Dataset and code can be found in \href{https://flytweety.github.io/AnimeDL2M/}{https://flytweety.github.io/AnimeDL2M/}.
\end{abstract}

\pagestyle{plain}
\maketitle
\vspace{20pt}

\section{Introduction}
The rapid advancements in AI-based image generation and editing methods, especially diffusion models \cite{rombach2022high}, have made image forgery increasingly accessible, sophisticated, and challenging to detect. Traditionally, image manipulations were primarily performed manually using tools like Photoshop \cite{coverage}. However, AI-based editing methods have significantly simplified the process \cite{bertazzini2024beyond}, resulting in highly realistic and difficult-to-detect forgeries \cite{ha2024organic}.

Although researchers have been aware of this threat and new datasets have been proposed, existing image manipulation detection and localization (IMDL) datasets and methods~\cite{sun2024rethinking, jia2023autosplice, trufor} are primarily tailored towards natural scenes and real-world photographs, neglecting domains such as anime imagery. Nonetheless, forged anime images are attracting increasing attention in areas such as copyright protection and content moderation \cite{nikkei2024animevsai}. Given their widespread popularity and extensive use across online communities and commercial markets, addressing forgery in anime images has become a crucial topic \cite{ha2024organic, nikkei2024aianime}. The absence of specialized forgery detection research in this domain represents a notable gap.

Unlike daily images, anime images have distinct visual characteristics such as unique color distributions, line patterns, texture styles, and structural details \cite{jin2025plagiarism}. Our experiment illustrates that existing forgery detection methods trained on daily images typically exhibit reduced performance in detecting and localizing forgeries within anime images. This limitation underscores the need for specialized datasets designed for IMDL tasks in anime domain.

To address these challenges, we introduce \textbf{AnimeDL-2M}, the first large-scale anime-specific image forgery dataset. In addition to its novel domain focus, AnimeDL-2M offers significant advantages in scale, generation variety, annotation richness, and content diversity. It comprises over 2 million images, including real, partially manipulated, and fully AI-generated samples. Fake images are created using six AI-based methods derived from three base models, ensuring both realism and variation and achieving high aesthetic quality scored state-of-the-art perceptual metrics. Each image is paired with comprehensive annotations, including image captions, objects, masks, mask labels and editing methods, enabling a broad range of downstream tasks. AnimeDL-2M also features rich diversity, with a broad set of object categories and manipulation scenarios, thereby providing a comprehensive benchmark for advancing research in AI-generated content detection.
\begin{table*}[t]
\small
\setlength{\tabcolsep}{12pt} 
\begin{tabular}{
>{\columncolor[HTML]{FFFFFF}}c 
>{\columncolor[HTML]{FFFFFF}}c 
>{\columncolor[HTML]{FFFFFF}}c 
>{\columncolor[HTML]{FFFFFF}}c 
>{\columncolor[HTML]{FFFFFF}}c 
>{\columncolor[HTML]{FFFFFF}}c }
\toprule
\cellcolor[HTML]{FFFFFF}{\color[HTML]{222222} }                          & \cellcolor[HTML]{FFFFFF}{\color[HTML]{222222} }                       & \multicolumn{2}{c}{\cellcolor[HTML]{FFFFFF}{\color[HTML]{222222} \textbf{\# Images}}} & \cellcolor[HTML]{FFFFFF}{\color[HTML]{222222} }                         & \cellcolor[HTML]{FFFFFF}{\color[HTML]{222222} }                                     \\ \cline{3-4}
\multirow{-2}{*}{\cellcolor[HTML]{FFFFFF}{\color[HTML]{222222} \textbf{Dataset}}} & \multirow{-2}{*}{\cellcolor[HTML]{FFFFFF}{\color[HTML]{222222} \textbf{Year}}} & {\color[HTML]{222222} \textbf{Real}}           & {\color[HTML]{222222} \textbf{Edited}}        & \multirow{-2}{*}{\cellcolor[HTML]{FFFFFF}{\color[HTML]{222222} \textbf{Domain}}} & \multirow{-2}{*}{\cellcolor[HTML]{FFFFFF}{\color[HTML]{222222} \textbf{Manipulation Types}}} \\ \hline
{\color[HTML]{222222} Columbia~\cite{columbia}}                                          & {\color[HTML]{222222} 2004}                                           & {\color[HTML]{222222} 183}            & {\color[HTML]{222222} 180}           & {\color[HTML]{222222} Daily}                                            & {\color[HTML]{222222} Random}                                                       \\ \hline
{\color[HTML]{222222} CASIAv1~\cite{casia}}                                           & {\color[HTML]{222222} 2013}                                           & {\color[HTML]{222222} 800}            & {\color[HTML]{222222} 921}           & {\color[HTML]{222222} Daily}                                            & {\color[HTML]{222222} Manual}                                                       \\ \hline
{\color[HTML]{222222} CASIAv2~\cite{casia}}                                           & {\color[HTML]{222222} 2013}                                           & {\color[HTML]{222222} 7,491}          & {\color[HTML]{222222} 5,123}         & {\color[HTML]{222222} Daily}                                            & {\color[HTML]{222222} Manual}                                                       \\ \hline
{\color[HTML]{222222} DSO-1~\cite{dso-1}}                                             & {\color[HTML]{222222} 2013}                                           & {\color[HTML]{222222} 100}            & {\color[HTML]{222222} 100}           & {\color[HTML]{222222} Daily}                                            & {\color[HTML]{222222} Manual}                                                       \\ \hline
{\color[HTML]{222222} Coverage~\cite{coverage}}                                          & {\color[HTML]{222222} 2016}                                           & {\color[HTML]{222222} 100}            & {\color[HTML]{222222} 100}           & {\color[HTML]{222222} Daily}                                            & {\color[HTML]{222222} Manual}                                                       \\ \hline
{\color[HTML]{222222} NIST16~\cite{nist}}                                            & {\color[HTML]{222222} 2016}                                           & {\color[HTML]{222222} 875}            & {\color[HTML]{222222} 564}           & {\color[HTML]{222222} Daily}                                            & {\color[HTML]{222222} Manual}                                                       \\ \hline
{\color[HTML]{222222} Fantastic Reality~\cite{fantastic}}                                 & {\color[HTML]{222222} 2019}                                           & {\color[HTML]{222222} 16,592}         & {\color[HTML]{222222} 19,423}        & {\color[HTML]{222222} Daily}                                            & {\color[HTML]{222222} Manual}                                                       \\ \hline
{\color[HTML]{222222} IMD20 Manual~\cite{imd2020}}                                      & {\color[HTML]{222222} 2020}                                           & {\color[HTML]{222222} -}              & {\color[HTML]{222222} 2,000}         & {\color[HTML]{222222} Internet}                                         & {\color[HTML]{222222} Unknown}                                                      \\ \hline
{\color[HTML]{222222} IMD20 Synthetic~\cite{imd2020}}                                   & {\color[HTML]{222222} 2020}                                           & {\color[HTML]{222222} -}              & {\color[HTML]{222222} 35,000}        & {\color[HTML]{222222} Daily}                                            & {\color[HTML]{222222} Random, Synthetic AI}                                         \\ \hline
{\color[HTML]{222222} tampered COCO~\cite{catnetv2}}                                     & {\color[HTML]{222222} 2022}                                           & {\color[HTML]{222222} -}              & {\color[HTML]{222222} 400,000}       & {\color[HTML]{222222} Daily}                                            & {\color[HTML]{222222} Random}                                                       \\ \hline
{\color[HTML]{222222} tampered RAISE~\cite{catnetv2}}                                    & {\color[HTML]{222222} 2022}                                           & {\color[HTML]{222222} 24,462}         & {\color[HTML]{222222} 400,000}       & {\color[HTML]{222222} Daily}                                            & {\color[HTML]{222222} Random}                                                       \\ \hline
{\color[HTML]{222222} COCOGlide~\cite{trufor}}                                         & {\color[HTML]{222222} 2022}                                           & {\color[HTML]{222222} -}              & {\color[HTML]{222222} 512}           & {\color[HTML]{222222} Daily}                                            & {\color[HTML]{222222} Synthetic AI}                                                 \\ \hline
{\color[HTML]{222222} AutoSplice~\cite{jia2023autosplice}}                                        & {\color[HTML]{222222} 2023}                                           & {\color[HTML]{222222} 2,273}          & {\color[HTML]{222222} 3,621}         & {\color[HTML]{222222} News}                                             & {\color[HTML]{222222} Synthetic AI}                                                 \\ \hline
{\color[HTML]{222222} MIML~\cite{miml}}                                              & {\color[HTML]{222222} 2024}                                           & {\color[HTML]{222222} -}              & {\color[HTML]{222222} 123,150}       & {\color[HTML]{222222} Internet}                                         & {\color[HTML]{222222} Unknown}                                                      \\ \hline
{\color[HTML]{222222} GRE~\cite{sun2024rethinking}}                                               & {\color[HTML]{222222} 2024}                                           & {\color[HTML]{222222} -}              & {\color[HTML]{222222} 228,650}       & {\color[HTML]{222222} Daily, News}                                      & {\color[HTML]{222222} Synthetic AI}                                                 \\ \hline
{\color[HTML]{222222} CIMD~\cite{cimd}}                                              & {\color[HTML]{222222} 2025}                                           & {\color[HTML]{222222} -}              & {\color[HTML]{222222} 600}           & {\color[HTML]{222222} Daily}                                            & {\color[HTML]{222222} Manual}                                                       \\ \hline
{\color[HTML]{222222} \textbf{AnimeDL-2M (Real \& Inpaint Subset)}}                                        & {\color[HTML]{222222} \textbf{2025}}                                           & {\color[HTML]{222222} \textbf{639,268}}        & {\color[HTML]{222222} \textbf{779,502}}       & {\color[HTML]{222222} \textbf{Anime}}                                            & {\color[HTML]{222222} \textbf{Synthetic AI}}                                                 \\ \bottomrule
\end{tabular}
\vspace{3mm}
\caption{Summary of public image IMDL datasets. AnimeDL-2M is the first IMDL dataset built with the latest diffusion models for anime images. Apart from real images and edited images, AnimeDL-2M also includes 884,129 fully AI-generated images.}
\label{tab:dataset}
\vspace{-5mm}
\end{table*}

Extensive experiments on AnimeDL-2M demonstrate a significant domain gap between the anime images and daily images. Considering the unique visual characteristics of anime, to better handle IMDL tasks on anime images, we propose \textbf{\modelname}, an IMDL model designed for anime images. It first employs a Mixed Feature Extractor to leverage texture information and object semantics in anime images. Then Dual-Perception Encoder is further introduced to encode and fuse texture-level cues with object-level semantics in two branches. Finally, feature maps are sent to Localization and Classification Predictor to get the prediction result. Through extensive comparative experiments, \modelname achieves superior performance compared to six leading SOTA models. 

Our main contributions are summarized as follows: \textbf{(1)} We introduce \textbf{AnimeDL-2M}, the first large-scale anime-specific IMDL dataset, featuring over 2 million images with rich annotations and high diversity. \textbf{(2)} We propose \textbf{\modelname}, a novel model tailored to synthetic anime detection with generalizability to in-the-wild images. \textbf{(3)} We demonstrate the domain gap between anime and daily images, which sheds light on future IMDL research. Our findings underscore the need for domain-specific solutions to support real-world applications such as copyright protection, content moderation, and intellectual property enforcement.

\section{Related Work}
This section reviews the studies in IMDL, with a focus on both datasets and model designs. We first summarize existing datasets, highlighting the lack of resources dedicated to anime imagery. We then introduce prior models, discussing their feature extraction strategies, backbone networks, and decoder designs, which offer design insights but also highlight the need for specialized solutions in the anime domain.

\subsection{IMDL Datasets}

Table~\ref{tab:dataset} summarizes the widely used datasets in IMDL research. Traditionally, most IMDL benchmarks~\cite{columbia, casia, coverage, nist, fantastic, imd2020, tampered} employ classical manipulation techniques such as copy-move, splicing, and object removal, which are often referred to as Photoshop-based methods~\cite{GIM}, only a limited number of datasets~\cite{trufor, MSCOCO, DEFACTO} include large-scale inpainting manipulations. With the rapid progress in generative modeling, text-guided image inpainting has become an emerging trend in dataset construction~\cite{GIM, tgif}. For instance, Jia et al.\cite{jia2023autosplice} proposed one of the earliest pipelines using DALL·E 2 to generate inpainted samples, while Sun et al.\cite{sun2024rethinking} expanded this approach by incorporating a wider range of generative models. In addition, several recent datasets~\cite{liu2024forgerygpt, fakeshield, forgerysleuth, sida, lian2024large, shao2024detecting} integrate auxiliary metadata or text annotations generated by large language models (LLMs), enabling the exploration of multimodal approaches and strengthening links to the broader domain of disinformation detection.

Despite these advances, existing datasets overwhelmingly focus on natural images, leaving anime-style content—an increasingly popular and distinct visual domain—largely underexplored. To address this gap, we introduce the first large-scale IMDL dataset specifically curated for anime imagery which reflects a real-world application scenario: the growing need for automated copyright protection and content integrity verification in AI-generated anime artworks. Our dataset is characterized by its rich annotations and high diversity, we anticipate that this contribution will stimulate further research at the intersection of multimedia forensics, generative media, and copyright governance.

\subsection{IMDL Models}
As summarized in~\cite{imdlbenco}, most existing models follow a common paradigm: First, they extract auxiliary features from the input image, then feed both the raw image and these features into an encoder network to obtain multi-scale feature maps. Finally, these features are fused and decoded to predict forgery locations and classification results.

Regarding input features, although some studies have demonstrated that auxiliary features are not strictly necessary~\cite{imlvit, SparseViT}, many state-of-the-art methods rely heavily on them. These include frequency- or edge-based representations extracted via handcrafted filters~\cite{objectformer, omni-iml} and noise-based features obtained through trained or learnable extractors~\cite{mvssnet, bayarconv, trufor, noiseprint, GIM, niu2024image, li2024noise, zhu2024learning}. Other studies incorporate semantics-aware features~\cite{chen2024leveraging, mesorch}, model-specific artifacts~\cite{npr, he2024rigid}, or compression-related cues such as JPEG artifacts~\cite{catnetv2, dualjpeg}. Some works further enhance detection by combining multiple auxiliary features~\cite{mmfusion, unionformer, karageorgiou2024fusion}.

In terms of encoder architectures, traditional approaches largely utilize CNN-based backbones, while more recent efforts have explored Transformer-based designs~\cite{imlvit, SparseViT, adaifl, guo2024effective, karageorgiou2024fusion, mgqformer} or hybrid architectures that combine both paradigms~\cite{objectformer, mesorch, unionformer}. Emerging directions also investigate the use of large vision encoders from LLMs~\cite{safire, su2024novel, li2024noise, yan2024sanity} or the integration of LLMs directly into the detection pipeline~\cite{fakeshield, liu2024forgerygpt, forgerysleuth, sida, hifinet2, lian2024large}. Additionally, effective fusion of diverse input features has become a key research focus~\cite{unionformer, mvssnet, karageorgiou2024fusion, zhang2024new, guo2024effective}.

Decoder designs have also evolved to better support localization tasks, where multi-scale feature maps play a critical role~\cite{mvssnet}. Various fusion strategies have been proposed to enhance feature aggregation~\cite{pscc, zhang2024new, guo2023hierarchical, yao2025dense, sumi-ifl, omni-iml}. In parallel, some studies focus on improving classification accuracy~\cite{trufor}. Moreover, contrastive learning has been employed to refine feature representation~\cite{zhou2023pre, unionformer, hao2024ec, safire, zhang2024image, bai2024image, liu2024attentive, lou2025exploring}, while other works introduce novel paradigms and frameworks for IMDL~\cite{wang2024hrgr, adaifl, forgeryttt, pan2024auto, liu2024multi, DH-GAN, yu2024diffforensics, zhang2024editguard, lou2024image}.

In this work, we conduct extensive experiments to construct a benchmark for AI-generated anime IMDL and propose the \modelname model. Leveraging the unique visual characteristics of anime-style imagery, our approach incorporates a hybrid feature extractor and a dual-branch architecture specifically designed to integrate texture-level cues with object-level semantic information. Extensive experiments demonstrate that this design achieves state-of-the-art performance in IMDL within the anime domain.

\section{AnimeDL-2M Dataset}

\begin{figure*}[tp]
    \centering
    \includegraphics[width=0.99\linewidth]{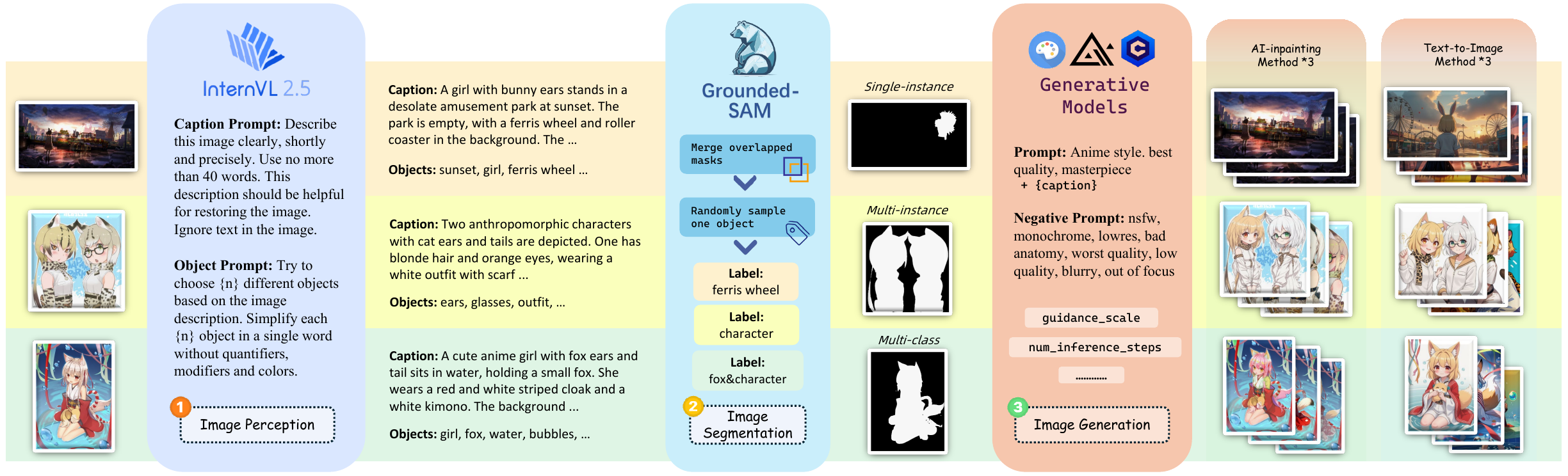}
    \vspace{-3mm}
    \caption{\small An overview of AnimeDL-2M's data construction pipeline and data example. Image perception component reads the image and outputs image caption as well as objects found in the image. Image segmentation component randomly picks one object and generates its mask for each image. Image generation component uses inpainting and text-to-image methods with 6 different models to create 6 fake images for each raw image. Captions, objects, mask labels, and editing methods serve as extra annotations.}
    \label{fig:dataset-pipeline}
\end{figure*}

This section presents AnimeDL-2M, the first million-scale IMDL dataset in the anime domain. We detail the dataset construction pipeline, including data collection, image perception, image segmentation, AI-based image generation, and dataset annotation, followed by an assessment of aesthetic quality and subject diversity. Compared to existing publicly available regional editing datasets detailed in Table \ref{tab:dataset}, AnimeDL-2M dataset offers significant advantages in scale, generation variety, annotation richness, and content diversity.
        
\subsection{Source Data Collection}
We collect raw images from Danbooru \cite{danbooru}, a widely-used art and anime platform that hosts high-quality, user-annotated images accompanied by detailed tags and textual descriptions. We resize the longer side of each image to 1024 pixels to achieve a balance between visual quality and computational efficiency. To evaluate the performance of benchmark models under realistic conditions, we additionally collected AI-generated anime images from Civitai \cite{civitai}, a popular community platform for AI-generated artwork contributed by users worldwide. Specifically, we retrieved the top 100 highest-rated anime generative models and filtered out those labeled with "realistic" tags or marked with high NSFW content. This filtering process yielded a curated set of 9,071 models spanning 14 base model categories, including Illustrious, Pony, Stable Diffusion (SD) 1.4/1.5/2.0/2.1, FLUX.1 S/D, SDXL 0.9/1.0/Hyper/Lightning/Turbo, among others. Utilizing the image URLs embedded in the metadata of these models, we collected a total of 104,627 high-quality text-to-image (T2I) samples, which serve as a challenging testset for evaluating our manipulation detection framework.

\subsection{Dataset Construction}
In order to simulate real-world image forgery scenarios while achieving a balance between efficiency and quality, we developed a fully automated pipeline based on large multi-modal models. This pipeline enables the creation of large-scale annotated datasets featuring diverse types of tampering content generated by multiple models.
As shown in Figure~\ref{fig:dataset-pipeline}, the pipeline consists of three key steps: (1) \textit{Image perception}, which generates a descriptive caption for each image; (2) \textit{Image segmentation}, which produces region masks to guide AI-based editing; and (3) \textit{Image generation}, which synthesizes both inpainted and fully AI-generated images.

\subsubsection{Image Perception}
In real-world scenarios, image manipulations are typically driven by specific intentions rather than being performed randomly or arbitrarily. Individuals must first understand the content of an image before making manipulations. In addition, manipulations often occur at the object level \cite{mesorch}, such as by removing, adding, or modifying particular objects. Based on these insights, the first stage of our pipeline aims to simulate the process of understanding and decision-making by leveraging a multimodal large model. This stage extracts critical information for downstream tasks. Specifically, we deploy the InternVL2.5 \cite{chen2024expanding}, which is one of the best open-source multi-modal large language model, to generate a concise description of the image, which we refer to as image caption. The model is then prompted to enumerate the objects present in the image, which will be used in Image Segmentation stage for generating masks. Given that the downstream image generation model employs CLIP’s text encoder \cite{radford2021learning}, which accepts a maximum of 77 tokens, we instruct the large model to produce captions that are both clear and succinct, constrained to fewer than 40 tokens. This leaves sufficient token capacity for additional input components required by the generation model.

\subsubsection{Image Segmentation}
After identifying the objects for manipulation, we uses them as labels to prompt the GroundedSAM \cite{ren2024grounded} to generate the mask for each object. To enhance the quality of the generated mask, we apply morphological closing operations to smooth its edges and reduce internal holes. During the mask generation process, a single label may yield multiple mask regions; in such cases, we merge them into a unified mask. Furthermore, if the Intersection-over-Union (IoU) between masks associated with different object labels exceeds 0.9, we treat them as overlapping representations and merge them into a single mask as well. As a result of the merging operations, we obtain three types of masks. The first and most common type is the single-instance mask, which contains a single instance of one object. The second type is the multi-instance mask, which includes multiple instances of the same object class. The third type is the multi-class mask, formed by merging highly overlapping masks from different object categories. The inclusion of diverse mask types further enriches the diversity of the AnimeDL-2M dataset and enhances its overall quality.

\subsubsection{Image Generation}
With the collected raw images, image captions, and object masks, we proceed to Text-to-Image synthesis and AI-inpainting image synthesis. The first step involves selecting appropriate generative models. Previous research has shown that different generative models may leave distinct fingerprints or artifacts in the generated images \cite{yang2023model}. To ensure diversity within the dataset and to construct a reliable benchmark for evaluation, it is crucial to employ a variety of generative models during the image synthesis process. Besides, given the anime-specific focus of our dataset, the selected generative models must be capable of producing high-quality images that faithfully adhere to the anime style. Following extensive screening and empirical evaluation, we selected three representative methods for each of the two generation tasks. After generation, we apply image evaluating model MPS \cite{zhang2024learning} to evaluate quality and filter out low quality images.

\subsubsection{Dataset Annotations}
It is worth noting that the intermediate information obtained from the first two stages of the data pipeline also constitutes valuable annotation data, which can be applied to broader evaluation and detection methodologies, such as further development of multimodal detection approaches based on text semantics or editing method attribution. 
Therefore, unlike IMDL datasets, we have additionally provided captions, objects, mask labels, and editing methods as extra annotations, which we anticipate to facilitate the future study.

\subsection{Dataset Statistics}
\begin{figure}
    \centering
    \includegraphics[width=0.7\linewidth]{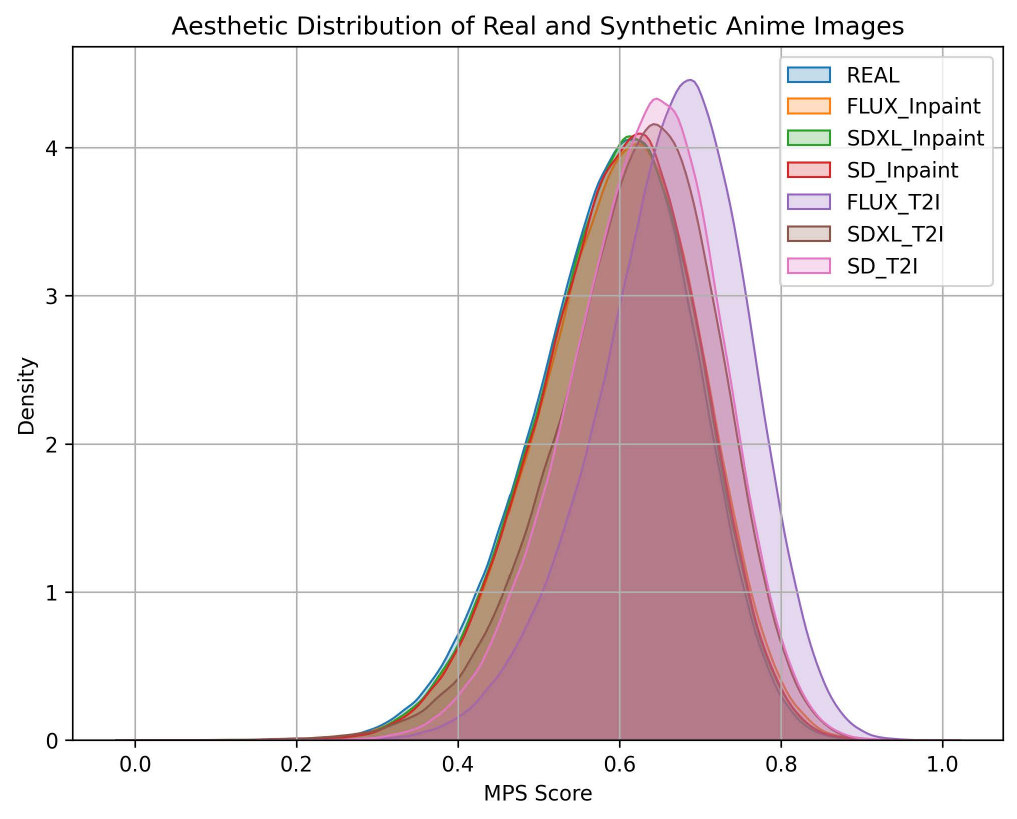}
    \vspace{-2mm}
    \caption{\small Aesthetic distribution of real and synthetic anime images. Note that inpainted images have a similar distribution to real images.}
    \label{fig:aesthetic}
    \vspace{-3mm}
\end{figure}

\begin{figure}
    \centering
    \includegraphics[width=0.8\linewidth]{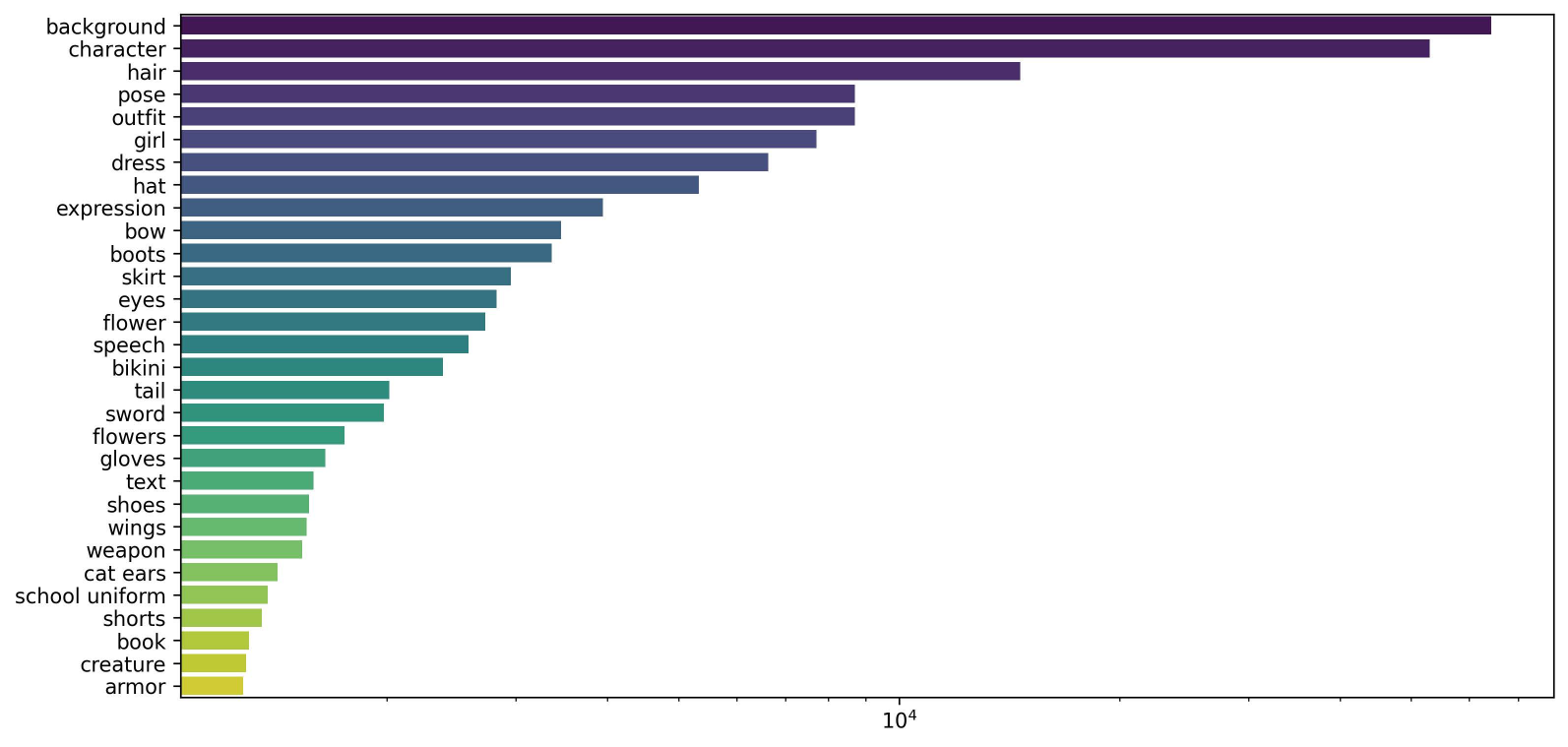}
    \caption{\small Top30 subject distribution of AnimeDL2M dataset. It exhibits a diverse range of subjects which highlights the open-world nature of the dataset, making it suitable for training robust and generalized IMDL models.}
    \vspace{-2mm}
    \label{fig:enter-label}
    \vspace{-2mm}
\end{figure}

Table~\ref{tab:animedl2m-statistics} summarizes the composition of the proposed AnimeDL-2M dataset. In Figure~\ref{fig:aesthetic}, We utilized MPS \cite{zhang2024learning}, a multi-dimensional preference scoring model for evaluating text-to-image generation, to access the image quaility of AnimeDL-2M dataset. We also present the most frequently manipulated subject when generating images with inpainting methods in Figure~\ref{fig:enter-label}. To summarize, AnimeDL-2M offers key advantages as below:

\textbf{New domain.} 
Anime images exhibit significant visual differences from daily images, resulting in a substantial domain gap. Models trained on existing daily image  datasets perform poorly when applied to anime images. As the first IMDL dataset in the anime domain, our dataset fills this critical gap and sheds light on many future research directions within the this field.

\textbf{Large-scale.} 
includes a large number of synthetic samples generated by different models under both Text-to-Image and AI-inpainting settings, as well as a substantial real image subset, outperforming most existing datasets and providing a comprehensive benchmark. Notably, the dataset is balanced across different generative methods and tasks. Its diverse content and balanced distribution benefit both evaluating and training IMDL models.

\textbf{Synthetic AI.} 
AI-generated image manipulations are becoming increasingly prevalent. Compared to traditional methods, AI-based edits often exhibit globally consistent styles and less distinguishable boundaries, making IMDL more challenging \cite{sun2024rethinking}. Unlike conventional IMDL datasets, AnimeDL-2M focuses on AI-based image manipulations and includes fully AI-generated images as well. Moreover, as shown in Figure~\ref{fig:aesthetic}, images in AnimeDL-2M generally receive high aesthetic scores, providing strong evidence of superior image quality and semantic consistency between images and annotations in the AnimeDL-2M dataset.

\textbf{Rich Annotation.}
For each group of original, edited, or generated images, AnimeDL-2M not only provides segmentation masks as in traditional datasets, but also includes additional annotations such as image captions, object descriptions, mask labels, and editing methods. These enriched annotations enable a broader range of tasks to be conducted on this dataset and are intended to facilitate future research in related domains.

{\renewcommand{\arraystretch}{0.6}
\begin{table}[t]
\centering
\small
\begin{tabular}{lll}
\toprule
\textbf{Subset}     & \textbf{Type} & \textbf{\# Images} \\ \midrule
Danbooru            & Real          & 639,268                   \\ \midrule
Stable Diffusion    & Text2Image    & 259,834                   \\ \midrule
Stable Diffusion XL & Text2Image    & 259,834                   \\ \midrule
FLUX                & Text2Image    & 259,834                   \\ \midrule
Stable Diffusion    & Inpainting    & 259,834                   \\ \midrule
Stable Diffusion XL & Inpainting    & 259,834                   \\ \midrule
FLUX                & Inpainting    & 259,834                   \\ \midrule
CivitAI             & Text2Image    & 104,627                   \\ \midrule
\textbf{Total}      & \textbf{-}    & \textbf{2,302,899}        \\ \bottomrule
\end{tabular}
\vspace{2mm}
\caption{\small Statistics of AnimeDL-2M Dataset}
\label{tab:animedl2m-statistics}
\vspace{-5mm}
\end{table}}

\textbf{High Diversity.}
AnimeDL-2M exhibits strong diversity across the following four dimensions: (1) three distinct types of segmentation masks; (2) six different image forgery methods based on three base models; (3) not only partially manipulated images, but also fully authentic and fully synthetic ones; and (4) diverse objects varying widely in type and content. As shown in Figure~\ref{fig:enter-label}, the distribution of manipulated subjects is fairly diverse and seemingly random, which contributes to model generalization and enables a more comprehensive evaluation of model performance.

\begin{figure*}
    \centering
    \includegraphics[width=0.88\linewidth]{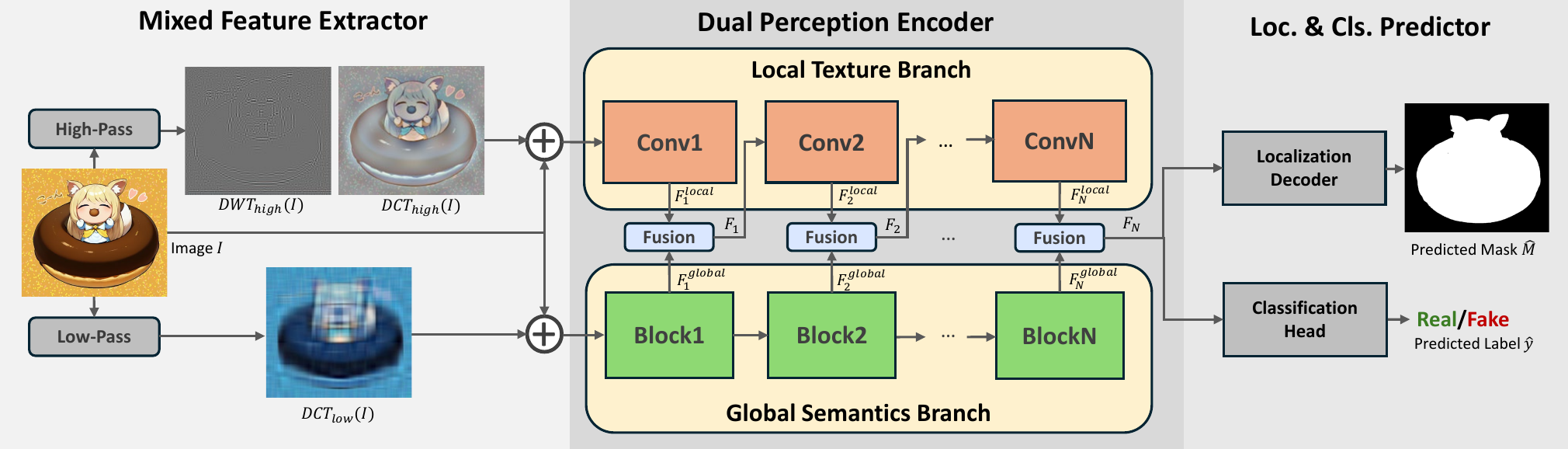}
    \caption{Overview of AniXPlore, which consists of Mixed Feature Extractor, Dual-Perception Encoder, and Localization and Classification Predictor, using information from both local textures and global semantics for anime IMDL.}
    \label{fig:anixplore}
    \vspace{-2mm}
\end{figure*}

\section{\modelname Model}
This section introduces \modelname, our proposed IMDL model tailored for the anime domain. We present the motivation behind the model design and introduce the overall architecture. Our model consists of a Mixed Feature Extractor, a Dual-Perception Encoder, and a Localization and Classification Predictor, aiming to capture forensic information from both local textures and global semantics.

\subsection{Inspiration and Design Overview}
Anime images exhibit distinctive visual characteristics that distinguish them from natural daily images, such as unrealistic lighting conditions, geometric abstractions, and the absence of sensor noise. These distinct properties underscore the necessity for specialized methods tailored to the IMDL tasks in the anime domain.

While it is commonly assumed that anime images contain fewer high-frequency components such as complex textures or stochastic noise, an overlooked yet crucial aspect is their retention of edge information in mid-to-high frequencies, especially the line contours. As anime images typically have clean and uncluttered scenes, line work in these images is generally sharp and well-defined. Furthermore, as these lines are manually drawn, they tend to exhibit a consistent artistic style across the image. Consequently, localized inconsistencies in stroke thickness, color, or drawing style may serve as effective cues for identifying image manipulations.

Additionally, prior studies have demonstrated that image manipulations frequently occur at the object level \cite{mesorch}. Anime images, which typically comprise a limited number of semantically salient objects with well-defined boundaries, are especially amenable to object-level semantic reasoning. Motivated by these insights, we propose \textbf{\modelname}, a model with dual-branch architecture that integrates semantic representations with frequency-aware features to enhance the IMDL in AI-generated anime images.

\subsection{Mixed Feature Extractor}
We integrate the Discrete Wavelet Transform (DWT) into the feature extraction pipeline to enhance high-frequency representation. DWT excels at preserving fine-grained edge structures, making it particularly effective in capturing line-based features such as contours and brush strokes, which serve as critical visual cues in anime images. 
Furthermore, inspired by \cite{mesorch}, which highlights the importance of object-level semantics in manipulation detection, we incorporate low-frequency components derived from the Discrete Cosine Transform (DCT). These features capture the global spatial structure of an image that are particularly relevant in identifying object-level manipulations. To be specific, the Mixed Feature Extractor combines DWT and DCT to process an RGB image $I \in \mathbb{R}^{3 \times H \times W}$. It computes 1) High-frequency components $M_{\text{high}}$, derived as the average of high-frequency DWT and DCT coefficients, and 2) Low-frequency DCT components $M_{\text{low}}$. The frequency components are concatenated with the original image $I$ to form mixed high-frequency input $M_{\text{high}}$ and mixed low-frequency input $M_{\text{low}}$ for the dual-branch encoder, where $M_{\text{high}}, M_{\text{low}} \in \mathbb{R}^{6 \times H \times W }$.
\begin{align}
M_{\text{high}} &= I \oplus \frac{1}{2} \left( \text{DWT}_{\text{high}}(I) + \text{DCT}_{\text{high}}(I) \right), \\
M_{\text{low}} &= I \oplus \text{DCT}_{\text{low}}(I),
\end{align}
where $\oplus$ denotes channel-wise concatenation.

\subsection{Dual-Perception Encoder}
To capture both localized textural patterns and global semantic information in anime images, we design a Dual-Perception Encoder comprising two complementary branches: a \textit{Local Texture Branch} and a \textit{Global Semantics Branch}. This dual-branch architecture ensures comprehensive feature extraction across multiple spatial scales and representation domains. The \textit{Local Texture Branch} is optimized to capture high-frequency details, which are particularly informative in the context of hand-drawn line art and forensic artifacts. We implement it using ConvNeXt \cite{liu2022convnet}, a state-of-the-art convolutional architecture that effectively models local patterns. The \textit{Global Semantics Branch} employs attention mechanisms to model long-range dependencies and contextual information. This branch facilitates semantic-level understanding, which is critical for detecting region-level inconsistencies and object-level manipulations. We implement it using Segformer \cite{xie2021segformer} for semantic feature extraction. At each encoding stage, we apply a $1 \times 1$ convolutional fusion layer to integrate features from both branches. The fused representation $F_i$ at stage $i$ is computed as:
\begin{equation}
F_i = \text{Fuse}(F_i^{\text{local}} \oplus F_i^{\text{global}}),
\end{equation}
where $F_i^{\text{local}}$ and $F_i^{\text{global}}$ are the outputs of the local and global branches at the $i$-th stage, respectively. The fused feature $F_i$ is then propagated to the subsequent layer of the local branch for progressive refinement with integrated local and global information. The encoder comprises 3 stages, each implementing the dual-branch extraction and fusion mechanism. The final fused output \( F_N \in \mathbb{R}^{384 \times \frac{H}{16} \times \frac{W}{16} } \) from the third stage is forwarded to the decoder.

\subsection{Localization and Classification Predictor}
Following \cite{imlvit}, we implement a Simple Feature Pyramid Network (SFPN) to transform the encoder output $F_N$ into multi-scale feature maps $\{F'_i\}_{i=1}^{5}$. Each $F'_i$ is resized to a uniform resolution of $ 256 \times \frac{H}{4} \times \frac{W}{4}$, after which the resized features are channel-wise concatenated and processed through a $1\times 1$ convolutional layer, yielding a fused feature map of dimensions $ 256 \times \frac{H}{4} \times \frac{W}{4}$. This fused representation is processed by a Multi-Layer Perceptron (MLP) to produce the predicted manipulation mask $\hat{M} \in \mathbb{R}^{1 \times \frac{H}{4} \times \frac{W}{4}}$, which is subsequently upsampled to the original resolution $H \times W$ to indicate potential forgery regions:
\begin{align}
\hat{M} = \text{MLP}\Bigl( \text{Conv}_{1\times1}\Bigl( \oplus_{i=1}^{5} \text{Resize}(F'_i) \Bigr) \Bigr), \space \space \{F'_i\}_{i=1}^{5} = \text{SFPN}(F_N)
\end{align}
For the classification head, we perform global max pooling on $F_N$ to yield a feature vector of shape $C\times1$, followed by a linear layer for binary prediction.
\begin{align}
\hat{y} = \text{Linear}(\text{MaxPool}(F_N))
\end{align}

\subsection{Loss Function}
We employ Binary Cross-Entropy (BCE) loss for both the localization and binary classification tasks.
Our experimental analysis indicates that incorporating a classification head can adversely affect localization performance. Recent works omit the classification head~\cite{mesorch} or use a two-stage training strategy~\cite{trufor, mmfusion}, both of which are either insufficient for comprehensive detection or unnecessarily complex. To address this issue, we employ an Automatic Weighted Loss (AWL) module inspired by the multi-task uncertainty weighting method~\cite{kendall2018multi} to mitigate the impact of loss imbalance. The overall loss is formulated as
\begin{align}
    \mathcal{L}_{\text{total}} = \frac{1}{2\sigma_1^2} \mathcal{L}_{\text{loc}} (M,\hat{M}) + \frac{1}{2\sigma_2^2} \mathcal{L}_{\text{cls}} (y,\hat{y}) + \log \sigma_1 + \log \sigma_2
\end{align}
where $M$ and $\hat{M}$ are ground-truth and predicted masks, $y$ and $\hat{y}$ are ground-truth and predicted labels. $\sigma _{1} $ and $\sigma _{2} $ are trainable parameters that represent the uncertainty of each task, allowing the model to adaptively adjust the relative importance of each loss component.

\section{Experiments}
This section describes our experimental setup and evaluation results. We benchmark the proposed \modelname model and the state-of-the-art methods on AnimeDL-2M and investigate domain gaps. We further examine generalizability through cross-dataset and in-the-wild evaluations, and perform ablation studies to assess the contribution of each design component in our model.

\subsection{Benchmark Settings}

\subsubsection{Baseline Models.} 
We selected six well-known, state-of-the-art open-source IMDL models from literature for comparative evaluation: Mesorch\_P (AAAI '25 \cite{mesorch}), MMFusion (MMM '24 \cite{mmfusion}), Trufor (CVPR '23 \cite{trufor}), IML-ViT ('23 \cite{imlvit}), CatNet (IJCV '22 \cite{catnetv2}), PSCC (TCSVT '22 \cite{pscc}), and MVSS (ICCV '21 \cite{mvssnet}). These models have demonstrated strong performance through their innovative design, and have been widely recognized by the research community, making them solid baselines for our experiments.

\subsubsection{Dataset Partition.}
We partitioned all image units (real images, synthetic images, and corresponding annotations) using an 8:1:1 train:validation:test ratio, excluding the Civitai subset which was reserved exclusively for evaluating cross-domain generalization. To facilitate fine-grained analysis of generative models' influence, according to the base models used for image generation, we further divide the training, validation, and test sets into three subsets: SD, SDXL, and FLUX. For evaluation purposes, we excluded fully authentic images (those without manipulated regions) since the F1 score metric becomes 0 in such cases. Additionally, we incorporated the GRE dataset \cite{sun2024rethinking}, a recent public benchmark containing over 200K AI-inpainted images, to investigate domain gaps between photographic and anime imagery. For the GRE dataset, we adhered to the partition scheme provided by the original authors.

\subsubsection{Experiment Tasks and Protocol.}
We designed four experimental tasks to systematically evaluate: (1) the efficacy of AnimeDL-2M and our proposed AniXplore; (2) the domain gaps between traditional and AI-based editing approaches, and between natural and anime images; and (3) the impact of architectural design on performance and generalization ability.

\textbf{Task 1 \& 2. Zero-shot and Training Results.} 
These two tasks evaluate the performance of pre-trained IMDL models on the AnimeDL-2M dataset, investigating the two types of domain gaps discussed earlier. In task 1, we initialize each model with checkpoints trained with Protocol-CAT \cite{imdlbenco}, a widely adopted protocol for IMDL evaluation. Three models are selected and further trained on the the GRE dataset, serving as representatives of IMDL models trained on IMDL datasets. In task 2, we use AnimeDL-2M dataset to both train each baseline model from scratch and finetune them using using the Protocol-CAT checkpoints, and compare them with \modelname trained on AnimeDL-2M.

\textbf{Task 3 \& 4. Cross-dataset and In-the-wild Tests.} 
These two tasks evaluate model performance on unseen data, providing insights into the model’s generalizability and inform future improvements. Specifically, we use the four models with classification head and initialized from checkpoints in the previous task to assess detection performance on the Civitai subset, serving as the in-the-wild test. Additionally, we retrain and evaluate both the detection and localization performance of three baseline models along with our \modelname on different subsets of the AnimeDL-2M to further investigate domain generalization. We report the score of the checkpoint that reaches the highest average pixel-level F1 score on all subsets.

\begin{table}[t]
\setlength{\tabcolsep}{4pt}
\small
\begin{tabular}{clcccc}
\toprule
\multirow{2}{*}{\textbf{Pretrain}} & \multirow{2}{*}{\textbf{Model}} & \multicolumn{2}{c}{\textbf{Pixel-level}} & \multicolumn{2}{c}{\textbf{Image-level}}
\\\cline{3-6}
 & & \textbf{F1} & \textbf{IoU} & \textbf{F1} & \textbf{Acc.} \\
\midrule
Protocol-CAT & MVSS~\cite{mvssnet}           & 0.0295        & 0.0169       & 0.8627         & 0.7587           \\
Protocol-CAT & PSCC~\cite{pscc}              & 0.0831        & 0.0553       & 0.4808         & 0.4008           \\
Protocol-CAT & CatNet~\cite{catnetv2}        & 0.0868        & 0.0561       & /              & /                \\
Protocol-CAT & TruFor~\cite{trufor}          & 0.0235        & 0.0161       & 0.8467         & 0.7408           \\
Protocol-CAT & MMFusion~\cite{mmfusion}      & 0.1106        & 0.0771       & 0.6941         & 0.5784           \\
Protocol-CAT & Mesorch\_P~\cite{mesorch}     & 0.0227        & 0.0151       & /              & /                \\
GRE          & MVSS~\cite{mvssnet}           & 0.0577        & 0.0322       & 0.8701         & 0.7702           \\
GRE          & CatNet~\cite{catnetv2}        & 0.2177        & 0.1419       & /              & /                \\
GRE          & Mesorch\_P~\cite{mesorch}     & 0.0655        & 0.0431       & /              & /                \\
\bottomrule
\end{tabular}
\vspace{2mm}
\caption{\small Zero-shot results on AnimeDL-2M. "Pretrain" denotes the dataset used for pretraining. GRE refers to \cite{sun2024rethinking}.}
\label{tab:zero-shot}
\vspace{-5mm}
\end{table}

\begin{table}[t]
\small
\begin{tabular}{lcccc}
\toprule
\multirow{2}{*}{\textbf{Model}} & \multicolumn{2}{c}{\textbf{Pixel-level}} & \multicolumn{2}{c}{\textbf{Image-level}}       \\\cline{2-5}
                                & \textbf{F1}        & \textbf{IoU}        & \textbf{F1}  & \textbf{Acc.} \\
\midrule
MVSS~\cite{mvssnet}           & 0.8802             & 0.8423        & 0.8991         &  0.8269            \\
PSCC~\cite{pscc}           & 0.9398             & 0.9106            & 0.9993     & 0.9989           \\
CatNet~\cite{catnetv2}         & 0.9447             & 0.9135              & /                 & /                 \\
TruFor~\cite{trufor}    &    0.9614     &     0.9362      &    0.9959   &    0.9938                  \\
MMFusion~\cite{mmfusion}     &   0.8615           &  0.8247       &    0.9581    &    0.9463   \\
Mesorch\_P~\cite{mesorch}    & 0.9661          & 0.9435            & /         & /  \\
\modelname            &  0.9710      &  0.9506                 &   0.9999       &    0.9998 \\
\modelname(HR)            &  0.9761      &  0.9923           &     0.9999       &    0.9999 \\
\bottomrule
\end{tabular}
\vspace{2mm}
\caption{\small Comparison of our model and existing SOTA IMDL models on  AnimeDL-2M.  Our model is trained from scratch. "HR" denotes high resolution version. For baseline models, we presents results that finetuned from Protocol-CAT checkpoint. Since pixel-level F1 score will be 0 for real images, pixel-level test only includes fake images while image-level test contains both real and fake images.}
\label{tab:finetune}
\vspace{-8mm}
\end{table}

\subsubsection{Metrics.}
We followed the same extensively used metrics for evaluation. For localization task, we use F1 score with throld = 0.5 and Intersection over Union (IoU). For detection task (for models with classification head), we use image-level F1 score and Accuracy.

\subsection{Implementation  Details}
We train \modelname on 8 H200 GPUs for 50 epochs with batch size of 72. All images were resized to $512\times512$ or padded to $1024\times1024$ pixels for two versions of \modelname. We used a cosine learning rate schedule, starting at 1e-4 and decaying to 5e-7, with a 2-epoch warm-up. The AdamW optimizer was applied with a weight decay of 0.05 to reduce overfitting. Gradient accumulation was set to 2 to effectively increase the batch size and enhance generalization. We use pre-trained backbones to initialize \modelname's two branches.

\begin{table*}[t]
\centering
\small
\setlength{\tabcolsep}{4pt} 
\begin{tabular}{clcccc|cccc}
\toprule
\textbf{Task} & \textbf{Method} & \textbf{Seen (FLUX)} & \textbf{Unseen (SD)} & \textbf{Unseen (SDXL)} & \textbf{Real}
& \textbf{Seen (SD)} & \textbf{Unseen (SDXL)} & \textbf{Unseen (FLUX)} & \textbf{Real} \\
\midrule
\multirow{5}{*}{Loc.} 
& MVSS 
& 0.9264 / 0.8893 & 0.0026 / 0.0015 & 0.0164 / 0.0129 & /
& 0.9298 / 0.8953 & 0.5581 / 0.5135 & 0.0852 / 0.0577 & / 
\\
& PSCC 
& 0.9254 / 0.8883 & 0.0009 / 0.0006 & 0.0031 / 0.0022 & /
& 0.9048 / 0.8684 & 0.0167 / 0.0108 & 0.0120 / 0.0074 & / 
\\ 
& Mesorch\_P 
& 0.9616 / 0.9362 & 0.0006 / 0.0005 & 0.0366 / 0.0339 & /
& 0.9609 / 0.9372 & 0.0644 / 0.0512 & 0.0210 / 0.0133 & / 
\\ 
& TruFor 
& 0.9621 / 0.9365 & 0.0015 / 0.0011 & 0.0520 / 0.0459 & / 
& 0.9617 / 0.9379 & 0.0627 / 0.0468 & 0.0388 / 0.0244 & /
\\ 
& \modelname 
& 0.9625 / 0.9375 & 0.0002 / 0.0001 & 0.0197 / 0.0175 & /
& 0.8940 / 0.8535 & 0.1767 / 0.1584 & 0.0315 / 0.0206 & / 
\\ 
& \modelname (HR) 
& 0.9774 / 0.9603 & 0.0000 / 0.0000 & 0.0005 / 0.0004 & /
& 0.9244 / 0.8841 & 0.1722 / 0.1198 & 0.0329 / 0.0213 & / 
\\ 

\midrule
\multirow{4}{*}{Det.} 
& MVSS 
& 0.5460 / 0.3755 & 0.1139 / 0.0604 & 0.1204 / 0.0641 & 0.9143
& 0.9984 / 0.9967 & 0.9442 / 0.8944 & 0.8811 / 0.7875 & 0.9950 
\\ 
& PSCC 
& 0.5038 / 0.3367 & 0.0082 / 0.0041 & 0.0101 / 0.0051 & 0.9974 
& 0.9984 / 0.9969 & 0.6824 / 0.5179 & 0.6702 / 0.5039 & 0.9963 
\\ 
& TruFor 
& 0.5468 / 0.3763 & 0.0206 / 0.0104 & 0.1211 / 0.0644 & 0.9997 
& 0.9977 / 0.9953 & 0.7490 / 0.5988 & 0.6586 / 0.4909 & 0.9968 
\\ 
& \modelname 
& 0.5107 / 0.3429 & 0.0021 / 0.0010 & 0.0283 / 0.0144 & 0.9997 
& 0.9989 / 0.9979 & 0.7922 / 0.6559 & 0.7147 / 0.5560 & 0.9534 
\\ 
& \modelname (HR)
& 0.5002 / 0.3335 & 0.0001 / 0.0001 & 0.0004 / 0.0002 & 1.0000
& 0.9999 / 0.9997 & 0.8629 / 0.7588 & 0.7169 / 0.5587 & 0.9922 
\\ 
\bottomrule
\end{tabular}
\vspace{1mm}
\caption{\small Cross-dataset evaluation results. Left columns show performance when trained on FLUX; right columns show performance when trained on SD. Metrics are F1 / IoU for localization and F1 / Accuracy for detection, and Accuracy for 'real' columns. HR denotes high resolution version.}
\label{tab:cross-dataset-merged-real} 
\vspace{-5mm}
\end{table*}

\subsection{Zero-Shot and Fine-Tuned Performance} 
\textbf{Domain Gaps.} As shown in Table \ref{tab:zero-shot}, the zero-shot performance of pretrained IMDL models on AnimeDL-2M is extremely poor. Most models achieve pixel-level F1 scores below 0.1, indicating a complete failure in localizing manipulated regions. This suggests that models trained on conventional IMDL datasets lack generalizability to the distribution of AI-edited anime images. Models pre-trained on the GRE dataset perform relatively better. This implies that certain features of AI-generated manipulations can be learned and partially transferred. However, these models still perform poorly on localization task. After fine-tuning on AnimeDL-2M, all models show significant improvements as reported in Table \ref{tab:finetune}, confirming both the high training value and the annotation quality of the dataset. Some models achieve surprisingly high F1 scores in detection task. This suggests that while models cannot precisely locate manipulated regions, they can still capture global statistical cues such as unnatural frequency artifacts or noise distribution that distinguish fake images from real ones at a coarse level. In the fine-tuned setting, all models also achieve relatively high localization scores. This is mainly because anime images tend to have clean backgrounds and less noise, which makes artifacts more visually distinct. These findings collectively validate the presence of substantial domain gaps across manipulation methods and image styles, especially for localization tasks. Therefore, AnimeDL-2M serves as a necessary contribution to bridge this gap, offering a dedicated benchmark for AI-edited anime image forensics.
 
\textbf{Comparison with SOTA.} As presented in Table \ref{tab:finetune}, our model achieves the best performance across all metrics. Although the absolute improvement over previous methods is modest, the gains are meaningful given that most baseline models already get very high scores. These results highlight the effectiveness of our model design in adapting to anime images and capturing manipulation artifacts specific to AI-generated content. Our approach offers beneficial insights for future research in anime and stylized media.

\subsection{Cross-Dataset and In-the-Wild Evaluation}

\indent \textbf{Factors Influencing Generalization.} As shown in Table~\ref{tab:cross-dataset-merged}, all models exhibit generally poor performance on the localization task under cross-dataset settings. This is likely because localization heavily relies on identifying artifact patterns left by AI-generated regions. However, such artifacts vary significantly across generative models, making it difficult to generalize. This also suggests that training or fine-tuning on the target domain can substantially improve localization performance, which is consistent with the findings in \cite{epstein2023online}. Combining results in Table~\ref{tab:civitai-test}, we can see that generalize ability on detection task does not strongly correlate with localization performance. Some models achieve high detection accuracy across domains despite limited localization ability. This implies that detection generalization may depend more on the robustness of the model architecture than on the ability to capture specific forgery artifacts. Among all tested methods, PSCC~\cite{pscc}, as the only model that uses RGB images as the sole input modality, demonstrates the weakest generalization, highlighting the importance of multi-modal or multi-channel feature inputs for generalizability. Meanwhile, both TruFor~\cite{trufor} and MMFusion~\cite{mmfusion} incorporate noise-based features, yet their performance differs significantly. This suggests that not all handcrafted features are equally effective, and the design of the feature extractor plays a critical role in mitigating overfitting. Therefore, careful selection and design of input features is essential for building more generalizable forensic models.

\begin{table}[h]
\small
\vspace{-1mm}
\begin{tabular}{lcc}
\toprule
                                 & \multicolumn{2}{c}{\textbf{Image-level}}                        \\ \cline{2-3} 
\multirow{-2}{*}{\textbf{Model}}          & \textbf{F1}                           & \textbf{Accuracy}   \\             
\midrule
MVSS \cite{mvssnet}                            & 0.9991                       & 0.9983                  \\
PSCC \cite{pscc}                            & 0.1233                       & 0.0657                  \\
TruFor \cite{trufor}        & 0.3908     & 0.2428      \\
MMFusion \cite{mmfusion} &          1.0000        &      1.0000
\\
\modelname                             & \textbf{1.0000} & \textbf{1.0000} \\
\bottomrule
\end{tabular}
\vspace{1mm}
\caption{\small Detection results on images collected from Civitai. Each models are trained on AnimeDL-2M.}
\label{tab:civitai-test}
\vspace{-7mm}
\end{table}

\textbf{Comparison with SOTA.} \modelname integrates both DWT and DCT as frequency-domain auxiliary features. This design enhances the model’s sensitivity to subtle traces and provides highly discriminative representations. \modelname achieves outstanding generalization in the detection task, obtaining perfect F1-score and accuracy in all sub datasets. These results demonstrate that \modelname can reliably identify fake images across a wide variety of generation models, which further confirms the robustness, versatility, and strong deployment potential of the proposed approach.

\subsection{Ablation Study}
Instead of examining designs that have been extensively validated in prior work, such as multiview feature maps or initializing the backbone with pretrained weights, we focus on evaluating the validity of three main components in \modelname. All experiments are conducted on the AnimeDL-2M dataset, with input images resized to 512×512. The results are presented in Table~\ref{tab:ablation}.

\textbf{Contribution of frequency features.} 
We observe a obvious performance improvement after introducing frequency-domain features, which demonstrates that our Mixed Feature Extractor effectively enhances the model’s perceptual ability.

\textbf{Feature fusion across latent levels.} We compare different strategies for feature fusion, including 1) Late Concat (LC): concat feature maps from all layers in two branches at once, 2) Progressive Concat (PC): fuse the feature map from each layer in two branches, and concat all fused feature maps, 3) Multiview Fusion (MF): fuse the feature map from each layer in two branches progressively and take fused feature from the last layer, then apply SFPN to obtain multiview feature maps from the fused feature. 
The results show that fusion between branches could be helpful, but fused feature maps may not always be the best feature map for decoder. It indicates that there may not be universally optimal fusion strategies, and the choice of fusion method should be tailored to the specific model architecture and task.
\begin{table}[h]
\setlength{\tabcolsep}{2pt}
\centering
\small
\begin{tabular}{ccccccc}
\toprule
\multirow{2}{*}{\textbf{Freq.}} & \multirow{2}{*}{\textbf{Fuse scheme}} & \multirow{2}{*}{\textbf{Task \& Loss}} & \multicolumn{2}{c}{\textbf{Pixel-level}} & \multicolumn{2}{c}{\textbf{Image-level}}
\\\cline{4-7}
& & & \textbf{F1} & \textbf{IoU} & \textbf{F1} & \textbf{Acc.} \\
\midrule
 --          & LC        &  Loc.       & 0.9456 & 0.9165 & --    & --    \\
 \checkmark & LC         & Loc.      & 0.9697 & 0.9485 & --    & --    \\
 \checkmark & LC        & Loc. + Cls. & 0.9633 & 0.9394 & 1.000 & 1.000 \\
 \checkmark & PC         & Loc.     & 0.9670 & 0.9447 & --    & --    \\
 \checkmark & MF & Loc. + Cls. + AW & \textbf{0.9710} & \textbf{0.9506} & \textbf{1.000} & \textbf{1.000} \\
\bottomrule
\end{tabular}
\vspace{1mm}
\caption{\small Ablation study on module-wise configurations.}
\label{tab:ablation}
\vspace{-7mm}
\end{table}

\textbf{Trade-off in multi-task learning.}
We found that directly introducing in classification head could cause a slight drop in localization performance, and the convergence of the overall loss becomes slower, suggesting a potential optimization conflict between the classification and localization objectives. Therefore, special measurements such as auto weight (AW) for loss should be taken into account when designing the loss function to achieve an optimal balance.

\section{Conclusion}

We present AnimeDL-2M, a large-scale dataset addressing the gap in IMDL datasets for anime images. Distinguished by its multiple generation methods, rich annotations, and high content diversity, AnimeDL-2M establishes a novel and expansive benchmark for anime-oriented IMDL tasks. Based on unique visual characteristics of anime images, we propose \modelname, a novel framework optimized for IMDL challenges in this domain. Our experiments reveal significant domain gaps between image styles and editing methods. Experimental results also show that \modelname outperforms existing SOTA methods in both detection and localization tasks on anime images, while exhibiting strong generalization capabilities in detection tasks. We aim to use AnimeDL-2M and \modelname to foster future innovations in this field.

\vspace{10mm}

\bibliographystyle{ACM-Reference-Format}
\bibliography{main}

\end{document}